\documentclass[runningheads]{llncs}

\usepackage{eccv}
\usepackage{eccvabbrv}
\usepackage{graphicx}
\usepackage{booktabs}
\usepackage[table]{xcolor}
\usepackage{colortbl}
\usepackage[accsupp]{axessibility}  %
\usepackage{hyperref}
\usepackage{multirow}
\usepackage{orcidlink}
\definecolor{blue-violet}{rgb}{0.54, 0.17, 0.89}

\newcommand{\method}{TacFiLM}
\newcommand\gderror[1]{
   \typeout{--------------------------------------------------------------------}
   \typeout{------- #1 ---------}
   \typeout{--------------------------------------------------------------------}
   {\bf #1}
}
\newcounter{gdTmp} 
\newcounter{gdLastCount}
\newcommand\maxpage[2][Error]{  %
\ifnum\value{page}>#2
    \gderror{On page {\thepage} we are past page #2 (too long).   #1 }
\else\fi
\setcounter{gdLastCount}{\value{page}} %
}
\newcommand\maxpageSinceLast[2][Error]{  %
\ifnum \numexpr \value{page} - \value{gdLastCount}\relax>#2
    \gderror{Exceeds max length #2 pages. Page \thepage: #1}
\thepage\else\fi
\setcounter{gdLastCount}{\value{page}} %
}

\begin{document}

\title{Tactile Modality Fusion for Vision-Language-Action Models} 

\titlerunning{\method{}}

\author{Charlotte Morissette \inst{1,2} \and
Amin Abyaneh \inst{1,2}\and
Wei-Di Chang \inst{1,2}\and
Anas Houssaini \inst{1,2} \and
David Meger \inst{1,2} \and
Hsiu-Chin Lin \inst{1,2} \and
Jonathan Tremblay \inst{3} \and
Gregory Dudek \inst{1,2}
}
\authorrunning{C. Morissette et al.}

\institute{
$^1$ McGill University, CAN \quad
$^2$ Mila - Québec AI Institute, CAN \quad
$^3$ NVIDIA, USA
}

\maketitle

\vspace{-19pt}
\begin{figure*}[h]
  \begin{center}
    \centerline{\includegraphics[width=\textwidth]{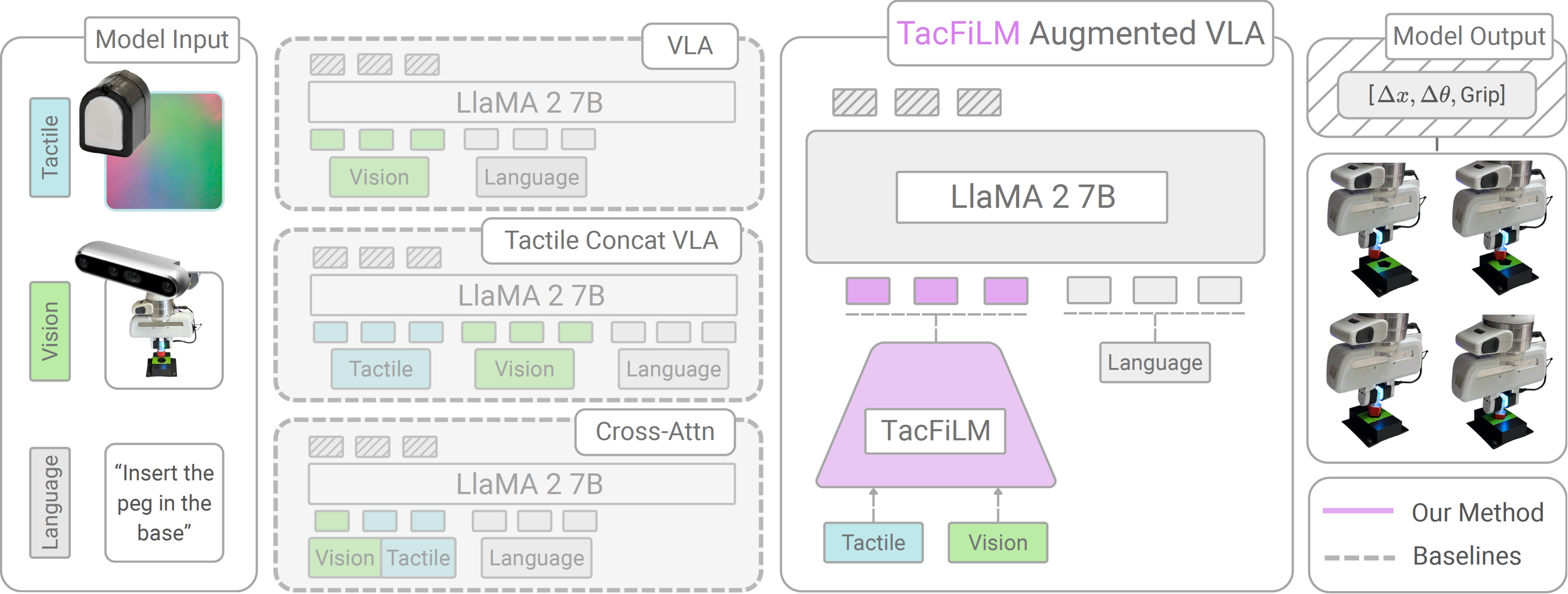}}
    \caption{
    \textbf{ \method{} Overview} We present \method{}, a lightweight modality-fusion approach for integrating visual-tactile signals into VLA models. The left panel shows the model inputs, including tactile, visual, and language modalities. In grey, baseline approaches. To the right, we show our proposed TacFiLM-augmented VLA. The rightmost boxes show model outputs and rollouts.
    }
    \label{intro_fig}
  \end{center}
\end{figure*}
\vspace{-35pt}

\begin{abstract}
\begin{sloppypar}
We propose \method{}, a lightweight modality-fusion approach that integrates visual-tactile signals into vision-language-action (VLA) models. While advances in VLAs have introduced robot policies that are both generalizable and semantically grounded, these models mainly rely on vision-based perception. Vision alone, however, cannot capture the complex interaction dynamics that occur during contact-rich manipulation, including contact forces, surface friction, compliance, and shear. While recent attempts to integrate tactile signals into VLA models often increase complexity through token concatenation or large-scale pretraining, the heavy computational demands of behaviour models necessitate lightweight fusion strategies. To address these challenges, \method{} outlines a post-training finetuning approach that conditions intermediate visual features on pretrained tactile representations using feature-wise linear modulation (FiLM). Experimental results on insertion and drawer opening tasks demonstrate consistent improvements in success rate, direct task performance, completion time, and force stability across both in-distribution and out-of-distribution tasks. Together, these results support our method as an effective approach to integrating tactile signals into VLA models, improving contact-rich manipulation behaviours.
\textbf{Project page: }\url{https://charliem7.github.io/projects/TacFilm/}
\end{sloppypar}
\end{abstract}
  
\section{Introduction}
\label{sec:intro}
In robotics, complex manipulation remains an open problem, with contact-rich tasks often exhibiting poor robustness to failure modes. Although existing approaches aimed at solving contact-rich tasks continue to rely predominantly on vision-based perception \cite{kim2024openvla, intelligence2025pi_, o2024open}, humans achieve remarkable dexterity by integrating visual, tactile, and proprioceptive sensory feedback \cite{johansson2009coding}. In particular, tactile signals provide critical information about object geometry, surface friction, and contact forces, complementing vision during physical interactions where occlusion and millimetre-scale adjustments limit visual feedback. Vision-based tactile sensors such as DIGIT \cite{lambeta2020digit} and GelSight \cite{yuan2017gelsight} capture this information as images, enabling integration with vision-based architectures.

\begin{sloppypar}
Recent work has begun integrating tactile signals into vision-language-action (VLA) models through finetuning or multimodal pretraining \cite{huang2025tactile, zhang2025vtla, cheng2025omnivtla, cheng2025touch100k}. These approaches face two challenges. First, VLA models require substantial data and compute for training and adaptation, creating demand for lightweight fusion strategies that remain accessible within a post-training finetuning paradigm. Second, the common baseline of feature concatenation often requires training separate tactile encoders and appends additional tokens to the VLA input, increasing sequence length and computational cost \cite{vaswani2017attention} while risking performance degradation as context grows \cite{wang2025mmlongbench, sharma2024losing}.
\end{sloppypar}

To tackle these problems, we propose \method{}, a novel lightweight modality fusion approach that integrates visual-tactile information into pretrained VLA models via feature-wise linear modulation (FiLM) \cite{perez2018film}, as seen in Figure~\ref{intro_fig}. We adopt a lightweight FiLM-based adaptation strategy rather than introducing additional cross-modal fusion modules, such as cross-attention \cite{heng2025vitacformer, zhao2025polytouch}, because it enables parameter-efficient adaptation without extensive multimodal pretraining. Our method leverages pretrained tactile representations and image conditioning, enabling tactile integration without increasing token sequence length or retraining large model components. In contrast to concatenation-based fusion strategies that append tactile embeddings to visual or language tokens, \method{} conditions intermediate visual features on tactile embeddings. This preserves pretrained visual-language priors while incorporating tactile signals.

\begin{sloppypar}
We evaluate our approach through real-robot experiments across over 1,000 rollouts on a diverse set of contact-rich insertion and pulling tasks, spanning both in-distribution and out-of-distribution settings. Our method consistently improves success rate, direct insertion/opening percentage, force stability and execution efficiency. Notably, for in-distribution tasks, our method achieves a 100\% success rate on the 3mm clearance Circle-Peg task and improves over the next-best baseline by up to 50\% on selected tasks. In the out-of-distribution setting, \method{} maintains strong performance with 100\% success rate on the 3mm clearance peg insertion tasks and improves HDMI cable plugging success by 30\%. It additionally reduces excessive interaction forces, requiring approximately one-third of the force applied by the baseline methods on select tasks. 
\end{sloppypar}

Overall, we show that FiLM-based fusion improves task performance while enhancing sensitivity to contact dynamics and reducing applied forces. Our main contributions are summarized as follows:

\begin{itemize}
    \item \method{}, a novel modality fusion approach that integrates tactile signals through image conditioning.
    \item Comprehensive experiments showing that \method{} improves success rates by up to 50\% with shorter episodes and reduced contact forces compared to concatenation and cross-attention-based fusion. 
    \item An investigation of the use of pretrained tactile encoders such as Sparsh \cite{higuera2024sparsh} and T3 \cite{zhao2024transferable} in fusing tactile signals into VLA models.
\end{itemize}

\section{Related Work}
\label{sec:related_work}

\subsection{Vision-Language-Action (VLA) Models}
Motivated by the success of large language models \cite{brown2020language, chowdhery2023palm, touvron2023llama, touvron2023llama2, wei2022chain} and vision-language models \cite{radford2021learning, li2022blip, li2023blip2, alayrac2022flamingo, chen2022pali, chen2023pali, jia2021scaling}, early approaches exploring the incorporation of semantic reasoning into robotics focused on high-level planning, often leaving action generation and execution to separate low-level controllers \cite{driess2023palm, ahn2022can, shah2023lm}. Others leveraged VLMs for robotics tasks such as affordance prediction, success detection, and representation learning, demonstrating improved semantic grounding and generalization across tasks \cite{du2023vision, shridhar2022cliport, karamcheti2023language, nair2022r3m, zhang2023grounding}. VLA models were subsequently introduced to unify semantic reasoning and action generation by leveraging semantic representations while directly grounding them in low-level robot control.
 
At a high level, VLA models aim to ground the semantic representations learned by vision-language models into physical action policies. These models typically combine a visual encoder, a projector, and a language model backbone to jointly process visual and textual tokens to generate output actions. Models such as OpenVLA \cite{kim2024openvla}, \(\pi_{0.5}\) \cite{intelligence2025pi_}, RT-1-X/RT-2-X \cite{o2024open}, and many others \cite{black2024pi_0, kim2025fine, team2024octo, brohan2022rt, zitkovich2023rt, stone2023open}, have been at the forefront of these advances.

Despite this progress, most existing VLA models primarily focus on vision and language modalities, with limited exploration of additional sensory inputs such as touch. This leaves open questions regarding how complementary modalities, particularly tactile signals, can be effectively incorporated into VLA frameworks for contact-rich manipulation. 

\subsection{Tactile Sensing in Robot Learning}
The development of vision-based tactile sensors such as GelSight \cite{yuan2017gelsight}, DIGIT \cite{lambeta2020digit}, and STS \cite{hogan2021seeing} has significantly advanced the integration of tactile perception in robotic manipulation. These sensors use an embedded camera to capture tactile information by recording the deformation of a gel membrane. Prior research has established tactile sensing as a complementary modality to vision in manipulation tasks, enhancing perception and control in contact-rich interactions \cite{blake2004neural, lee2019making}. Their high spatial resolution allows these sensors to capture slip, contact geometry, and deformation with high fidelity \cite{yuan2017gelsight, lambeta2020digit, hogan2021seeing}. This has made them a natural choice for integration into robotic end-effectors. They are often used to solve contact-rich robotic tasks such as insertions, grasping, and in-hand manipulation, where tactile feedback provides critical information about physical interactions \cite{lambeta2020digit, yuan2024robot, hogan2018tactile, wilson2023cable, calandra2018more, dong2021tactile, she2021cable, qi2023general, hansen2022visuotactile, cui2025vi}.  

Prior work has begun exploring the integration of tactile signals into foundation models, such as VLA models. At a high level, these approaches can be divided into two categories: those that incorporate tactile information during finetuning, and those that rely on additional multimodal pretraining to align tactile representations. The post-training finetuning approaches introduce tactile inputs during model finetuning, often by encoding tactile signals as additional tokens that are concatenated with the original VLM inputs. These methods rely on attention-based fusion during action generation \cite{huang2025tactile, zhang2025vtla, li2025adaptive, yu2024octopi, hao2025tla, bi2025vla, yu2025forcevla}. Conversely, the other category of approaches focuses on learning shared vision-tactile representations through large-scale pretraining or contrastive learning to align modalities prior to policy learning \cite{cheng2025omnivtla, cheng2025touch100k, yang2024binding, george2025vital, jones2025beyond, heng2025vitacformer, zhao2025polytouch, zhang2026touchguide}. However, as large behaviour models demand increasing amounts of data and compute for training and finetuning, we are motivated to explore lightweight modality fusion strategies within the post-training finetuning paradigm.

Within tactile fusion approaches, the most prevalent methods are concatenation and cross-attention. Concatenation-based methods integrate tactile embeddings by appending them to visual and language tokens, enabling the transformer to reason across all modalities \cite{zhang2025vtla, hao2025tla, yu2024octopi, yu2025forcevla}. Cross-attention-based methods instead allow visual and tactile representations to attend to one another through dedicated attention layers \cite{heng2025vitacformer, zhao2025polytouch}. Although these approaches have demonstrated promising results, their implementation either increases the token sequence length or requires training additional attention parameters, both increasing computational overhead. Our approach, in contrast, proposes a lightweight fusion strategy that conditions visual features on tactile information, maintains token lengths, and requires minimal additional training.

To support post-training finetuning strategies, pretrained tactile representations are important. Similar to pretrained visual backbones, these models learn tactile embeddings that encode tactile features such as contact, geometry, and deformation. Methods such as Sparsh \cite{higuera2024sparsh}, T3 \cite{zhao2024transferable}, and others \cite{gupta2025sensor, feng2025anytouch, feng2026anytouch, yang2024binding, ma2025cltp} have begun to explore self-supervised and large-scale pretraining approaches for learning transferable tactile representations.

Our approach builds on these advances by investigating how pretrained tactile representations can be effectively integrated into VLA policies within a post-training finetuning method. Our method further addresses limitations of naive fusion approaches, which often require training additional tactile encoders and increasing token sequence length.

\section{Methodology}
\label{sec:methodology}
\method{} outlines a tactile modality fusion approach that conditions VLAs on visuotactile representations. Unlike prior work, we aim for tactile integration without increasing the number of input tokens or requiring task-specific encoders. The former is achieved by employing a lightweight conditioning strategy as explained in Section~\ref{sec:policy_arch}, while the latter is the direct result of using pretrained tactile representations as seen in Section~\ref{sec:tactile_rep}. Consistent with the design philosophy of generalist VLAs, our fusion strategy of pretrained tactile representations in \method{} reduces the need for extensive retraining and repeated sensor-specific data collection.

\subsection{Policy Architecture}
\label{sec:policy_arch}
Figure \ref{vla-architecture} illustrates the overall architecture of our tactile-conditioned VLA policy. The model builds upon the OpenVLA-OFT framework \cite{kim2025fine}, which combines a fused SigLIP \cite{zhai2023sigmoid} and DINOv2 \cite{oquab2023dinov2} visual backbone, a lightweight MLP projector, and a decoder-only Llama2 7B language model \cite{touvron2023llama2}. To leverage tactile signals effectively within the VLA policy, we propose a modality fusion approach that integrates pretrained tactile embeddings into the VLA backbone. In contrast to approaches that jointly train modality encoders with the policy \cite{hao2025tla, huang2025tactile, yu2025forcevla, zhang2025vtla}, \method{} leverages pretrained tactile representations and aims to preserve the visual-language priors learned by the original VLA model. Specifically, we propose conditioning visual features with tactile information through a FiLM-based fusion approach.

At each time step, the image input and language prompts are processed following the standard VLA pipeline. Input images are encoded into patch-level visual embeddings by the fused vision backbone. In parallel, tactile observations are encoded using a pretrained tactile representation model. This tactile embedding is then used to condition intermediate vision representations. 

The resulting tactile-conditioned visual features are projected into the language model input space and concatenated with text tokens before being processed by the decoder-only LLM. The decoder's final hidden states are then passed to an MLP action head, which directly regresses continuous robot actions using an L1 regression objective. This design enables the model to jointly reason over visual, tactile, and language information while predicting continuous action chunks for robotic manipulation.

\subsubsection{FiLM-Based Fusion}
Feature-wise Linear Modulation conditions a backbone on an auxiliary signal by learning per-channel scale and shift \((\gamma\),\(\beta\)) parameters from the new input and applying a feature-wise affine transformation to intermediate activations \cite{perez2018film}. This auxiliary signal is a visuotactile image encoded with a pretrained tactile representation discussed above. At each time step \(t\), tactile images are encoded, and the resulting patch features are averaged to obtain a pooled tactile embedding \(z_t\). For each selected ViT block \(n\) in the VLAs visual encoders, DINOv2 and SigLIP, an MLP projects \(z_t\) to \(\gamma_n\) and \(\beta_n\) FiLM parameters. We apply FiLM after normalization and before multi-head self-attention as a feature-wise affine modulation of the intermediate visual features \(F_n\) as shown in the following equation: 

\begin{equation}
    \label{eq:FiLM}
    \text{FiLM}(F_n|\gamma_n, \beta_n) = F_n \odot (1 + \gamma_n) + \beta_n. 
\end{equation}

\begin{figure*}[t]
  \begin{center}
    \centerline{\includegraphics[width=\textwidth]{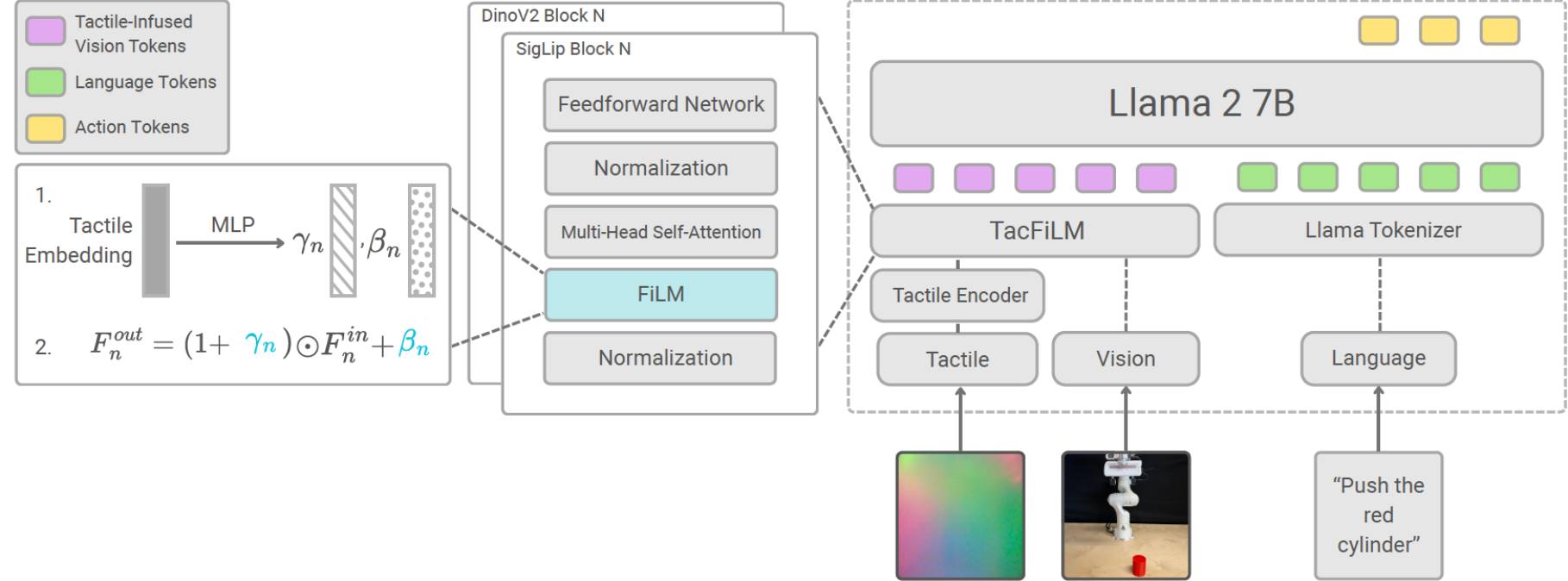}}
    \caption{\textbf{\method{}'s modality fusion pipeline.}  Tactile embeddings are integrated into the vision backbone immediately preceding the multi-head attention layers. The resulting multimodal tokens, combined with language inputs, serve as the basis for action generation within the Llama backbone.}
    \label{vla-architecture}
  \end{center}
\end{figure*}

The model architecture is shown in Figure \ref{vla-architecture}. Following the design principles from OpenVLA-OFT, \(\gamma\) and \(\beta\) are applied to the entire feature map. We initialize \(\gamma\) and \(\beta\) to zero, so conditioning starts near identity. We opt for a FiLM approach because it is computationally lightweight, provides a low-dimensional, inspectable global tactile bias, eliminates the need to append additional tokens to the language backbone, and integrates cleanly into ViT blocks. By default, \method{} applies FiLM conditioning to all ViT blocks in the visual encoder; we ablate this choice in Section~\ref{sec:ablation}.

\subsection{Pretrained Tactile Representations}
\label{sec:tactile_rep}
Our fusion approach is agnostic to the choice of tactile encoder. We evaluate two architectures with different pretraining objectives, T3 and Sparsh, to validate this flexibility. Both produce fixed-dimensional embeddings compatible with our FiLM conditioning mechanism.

\subsubsection{T3}
The T3 model \cite{zhao2024transferable} is a tactile representation framework that extends to various visuotactile sensors and multiple downstream tasks. The architecture includes sensor-specific encoders, task-specific decoders, and a shared transformer trunk. Each sensor is associated with its own independent encoder, while downstream tasks are managed by specific decoders, enabling transfer across sensors and tasks. For our application, we retain only the sensor encoder and shared transformer trunk. Each sensor encoder is implemented using a Vision Transformer (ViT) \cite{dosovitskiy2020image}, which processes tactile images as patch tokens. The encoder maps raw tactile observations into latent feature representations. These features are subsequently processed by a shared ViT-based transformer trunk, which refines the embeddings into unified tactile representations. All tactile frames are resized to 224×224 before being processed by the encoder. 

\subsubsection{Sparsh}
The Sparsh model \cite{higuera2024sparsh} is a pretrained tactile representation framework that learns generalizable features from large-scale tactile data. This method adopts a ViT architecture \cite{dosovitskiy2020image} trained using various self-supervised learning (SSL) objectives. Sparsh processes tactile images through a ViT encoder, where image inputs are tokenized into patch embeddings and passed through transformer blocks to produce latent tactile representations. Different Sparsh variants correspond to the three SSL paradigms they propose: Sparsh-MAE, Sparsh-IJEPA and Sparsh-DINO. We investigate all three. \textbf{Sparsh-MAE} uses a masked autoencoding objective where an encoder learns contextual representations of masked images, enabling a decoder to reconstruct the masked regions. \textbf{Sparsh-IJEPA} uses a joint-embedding predictive objective, learning representations by predicting latent features of masked regions rather than reconstructing raw inputs. \textbf{Sparsh-DINO} uses a self-distillation objective, where a student network learns to predict the representations of a teacher network. For all Sparsh variants, image preprocessing is identical. Two tactile frames, separated by five time steps, are concatenated channel-wise. The background is removed, and the resulting images are resized to 224×224.

\subsection{Training}
Despite large-scale pretraining, off-the-shelf VLAs often lack the precision required for specialized tasks. This motivates the use of post-training, specifically efficient finetuning recipes such as Low-Rank Adaptation (LoRA) \cite{hu2022lora}, to adapt the generalist model to specific domains without losing its pre-trained representations. We build on the observation that parameter-efficient finetuning provides an effective mechanism for adapting generalist VLAs, and extend this approach to integrate tactile signals into these models. We LoRA-finetune the linear layers of the \method{}-augmented VLA, namely, the OpenVLA-OFT and tactile backbone linear layers, while training the model's FiLM layers from scratch. By freezing most of the model and updating only a targeted subset of weights, we inject novel visuotactile modalities into the policy. In doing so, we aim to retain the base VLA's semantic understanding while leveraging rich tactile feedback for contact-rich manipulation.

\section{Experiments}
We conduct a series of experiments to evaluate the proposed modality fusion approach and the use of pretrained tactile representations in large pretrained models. Our experiments are guided by the following research questions: \textbf{Q1:} How do different pretrained tactile encoders (T3, Sparsh variants) influence downstream policy performance? \textbf{Q2:} How do different modality fusion mechanisms compare in integrating tactile signals into VLA models? \textbf{Q3:} Does the proposed \method{} improve task success, execution efficiency, and contact sensitivity?

\subsection{Experiment Setup}
\label{sec:exp_setup}

\begin{figure}[t]
  \begin{center}
    \centerline{\includegraphics[width=\linewidth]{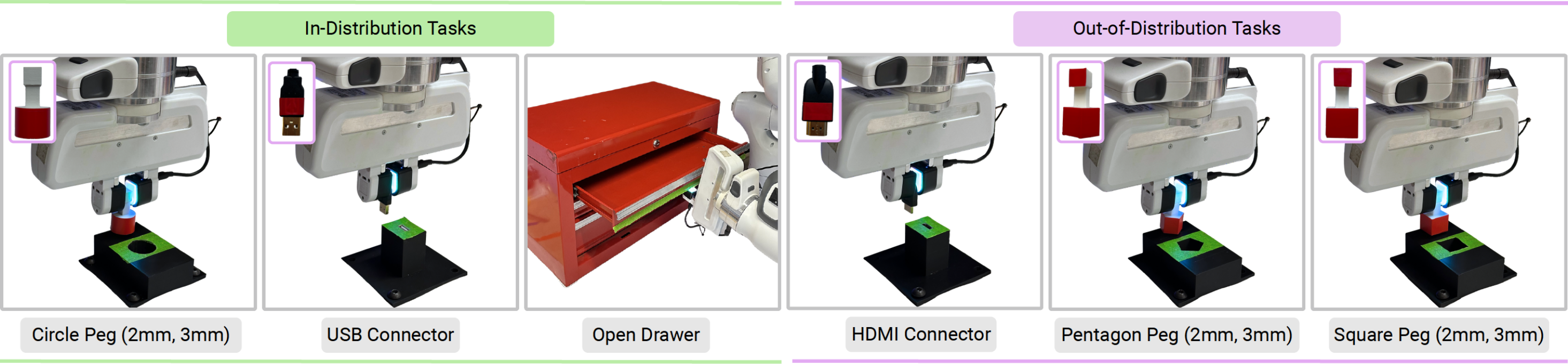}}
    \caption{\textbf{Task definitions.} Insertion tasks differ in peg or connector shape and clearance but share the goal of successful insertion. Open-drawer consists of hooking the gripper under the drawer and pulling it open.
    }
    \label{fig:task_setup}
  \end{center}
\end{figure}

\subsubsection{Tasks and Data Collection}
Our experimental evaluation focuses on a diverse suite of insertion and pulling tasks, depicted in \Cref{fig:task_setup}, and designed to test contact-rich manipulation capabilities. The benchmark of peg insertions and cable plugging spans multiple object geometries, varying peg and connector shapes, and introduces different levels of difficulty by varying insertion clearances. These variations are important because they create an environment in which direct insertions are difficult and the policy must often rely on tactile feedback for fine-grained adjustments. The benchmark also includes a drawer-opening task, extending the evaluation beyond insertion to further contact-rich scenes.

In each task, the objective is strictly defined: the robot must either fully insert the object into the target base or fully open the drawer, whether directly or through one or more corrective adjustments. We do not evaluate the model on its grasping abilities, and as a result, the robot begins all trajectories with the object in hand for insertion tasks. 

To evaluate our approach on a real-world setup, we collected expert demonstrations by teleoperating a Franka Emika Panda arm to complete the previously defined tasks. High-level robot control and teleoperation were implemented via Polymetis\footnote{\url{facebookresearch.github.io/fairo/polymetis}}, while low-level commands were transmitted through libfranka over the Franka Control Interface (FCI) at a control frequency of 1 kHz. The FCI provided real-time feedback of the robot's state to the controller. An operator controlled the end-effector pose using a 3Dconnexion SpaceMouse, whose 6-DoF inputs were mapped to Cartesian pose commands with tunable gains. For each task, we collected 80 demonstrations, each of approximately 70 steps. During data collection, we recorded time-aligned robot observations at 10 Hz, including joint positions and velocities, end-effector pose, gripper width and status, RGB and tactile images from an Intel RealSense camera and a DIGIT sensor, respectively, and executed actions. A fixed natural language description was appended to each demonstration. For insertion tasks, prompts followed the template: ``Insert the [colour] [shape] peg into the [colour] base''. For the drawer-opening task, the prompt was the following: ``Hook the gripper under the green handle of the top drawer and pull it open''.

\subsubsection{Evaluation Methods}
\begin{sloppypar}
To address the research questions outlined above, we compare \method{} against the following baseline methods. 
1) \textbf{OpenVLA-OFT:} OpenVLA-OFT is the base VLA model upon which all other approaches are built. The model combines a fused SigLIP \cite{zhai2023sigmoid} and DINOv2 \cite{oquab2023dinov2} visual encoder, an MLP projector and a decoder-only Llama2 7B \cite{touvron2023llama2} backbone. It is trained on visual observations and language prompts, serving as a vision-only baseline with which to compare our approach.
2) \textbf{TactileConcat:} We also consider feature concatenation, as implemented in prior work \cite{huang2025tactile, zhang2025vtla, li2025adaptive, yu2024octopi, hao2025tla, bi2025vla}. In our implementation, tactile images are embedded using the pretrained T3 or Sparsh models described above. The resulting tactile features are then projected to the VLM’s input space with a learned two-layer MLP. The resulting tactile tokens are concatenated with the VLM language and image tokens, which are then passed as input to the model. 
3) \textbf{Cross-Attn:} Lastly, we implement a cross-attention-based fusion strategy as proposed in prior work \cite{heng2025vitacformer, zhao2025polytouch}. Our implementation follows PolyTouch’s architecture \cite{zhao2025polytouch}. Tactile images are first embedded using the pretrained T3 or Sparsh models and projected into the visual feature space. Cross-attention is then applied after the vision backbone through six stacked cross-attention blocks with residual connections, in which visual patch embeddings serve as queries that attend to the projected tactile embeddings as keys/values. 

To compare our approach against the tactile fusion mechanisms proposed in prior work, we implement each baseline on a shared backbone, OpenVLA-OFT \cite{kim2025fine}, so that differences in performance can be attributed to the fusion strategy rather than the underlying architecture.

We further evaluate the effects of FiLM integration location through an ablation study, proposing variants of our model, \method{}, in which FiLM is integrated only into a subset of ViT blocks.
\end{sloppypar}

\begin{figure}[t]
  \begin{center}
    \centerline{\includegraphics[width=\linewidth]{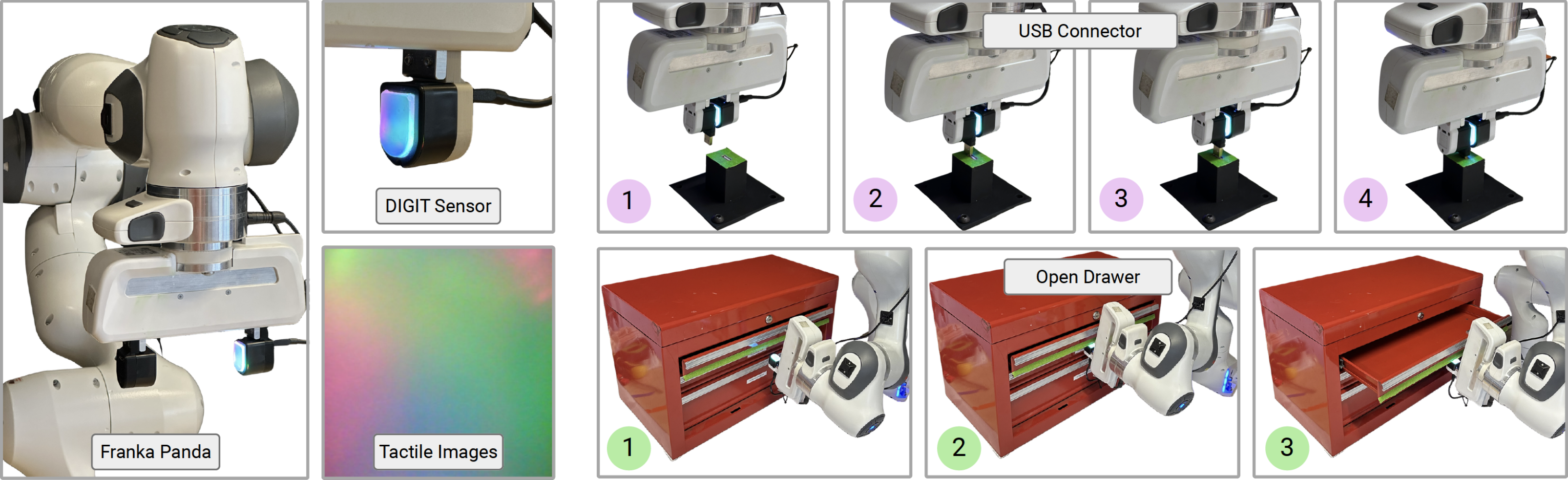}}
    \caption{\textbf{Experiment setup (left) and sample rollouts (right).} Franka parallel grippers, which the robot uses to hold the pegs, have a DIGIT tactile sensor installed on them. The rollouts demonstrate USB connector insertions and open-drawer tasks.
    }
    \label{fig:setup_rollouts}
  \end{center}
\end{figure}

\subsubsection{Evaluation Metrics}
To evaluate the performance of the methods listed above on our tasks, we compare task success rate (\%), percentage of direct insertions/openings, average maximum force exerted (N) and average time to task completion across all rollouts (s). These metrics were selected to capture different aspects of manipulation performance. Success rate reflects overall task reliability, including both direct and recovered behaviour, while the percentage of direct insertions/openings measures the model’s ability to achieve accurate alignment and grip without resorting to recovery behaviours. The average maximum force allows us to assess whether the policy accounts for contact dynamics during task execution. Finally, average task completion time reflects execution efficiency. We report mean values across trials to characterize overall expected performance. All deployed methods were trained to 80k steps. A task is deemed successful if the object in hand is fully inserted into the base or the drawer is fully opened. We distinguish direct insertions/opening (first attempt) from recovered insertions/openings (after adjustment).

\subsubsection{Hardware}
We run all deployments on the Franka Emika Panda\footnote{\url{https://franka.de/documents}}, a robot arm with 7 degrees-of-freedom. The arm is equipped with a parallel two-finger gripper, the Franka hand, on which is mounted a DIGIT visuotactile sensor, as seen in Figure \ref{fig:setup_rollouts}.

\subsection{Experimental Results}
We evaluate method performance on both in-distribution and out-of-distribution tasks to assess task-specific performance and generalization capabilities. In total, we conduct over 1,000 rollouts: 480 for in-distribution evaluation (30 per method), 300 for out-of-distribution (15 per method), and an additional 240 for ablation studies.

\begin{figure}[t]
  \begin{center}
    \centerline{\includegraphics[width=\textwidth]{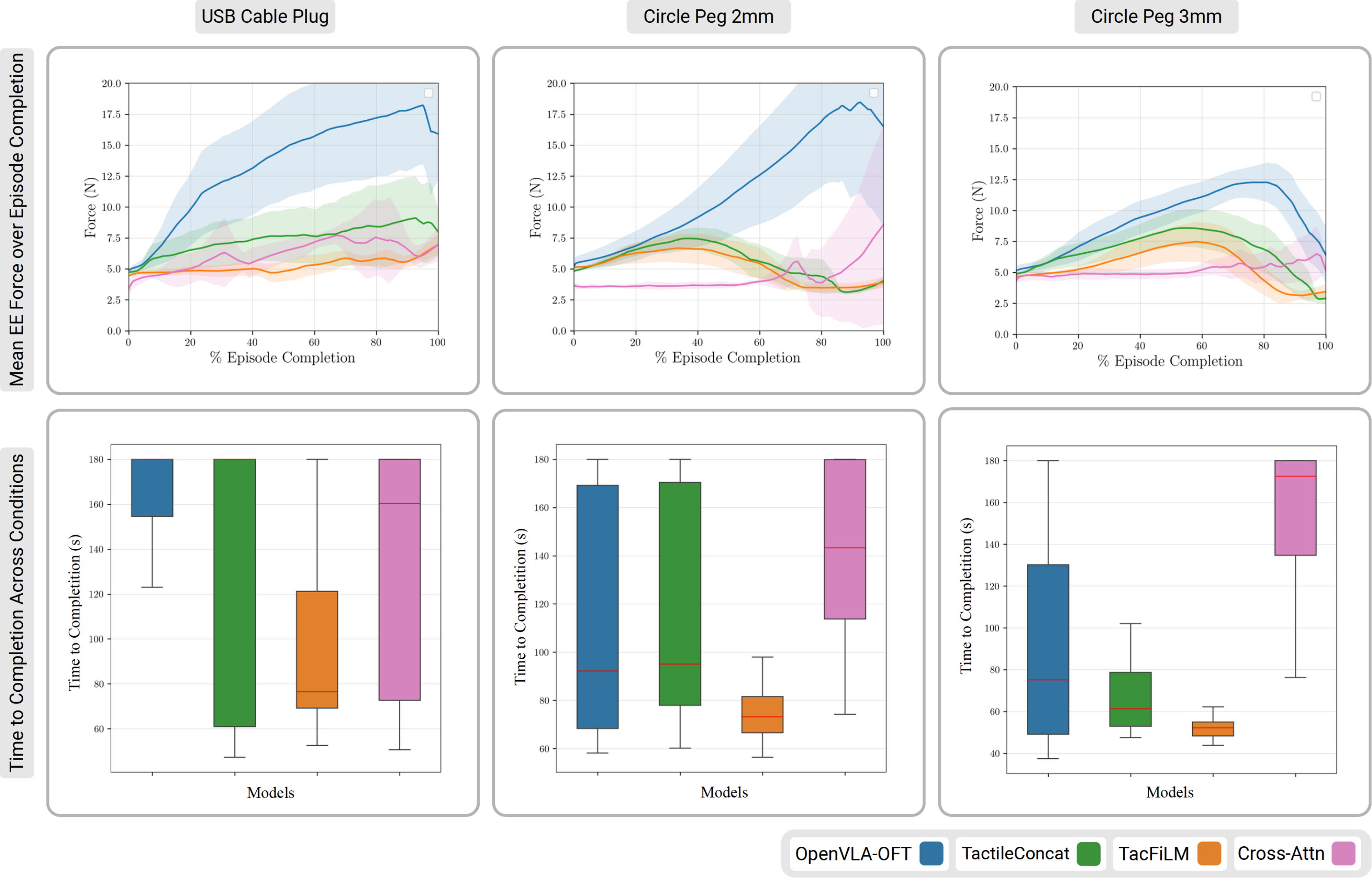}}
    \caption{\textbf{Force and task completion time analysis.} Top row: Average force measurements across successfully recovered insertions for the in-distribution (ID) tasks. Bottom row: Task completion times across different methods for ID insertion tasks. The results demonstrate that tactile-aware methods prevent excessive force application while \method{} also significantly reduces task completion time.}
    \label{fig:main-plots}
  \end{center}
\end{figure}

\subsubsection{In-Distribution Tasks}
We compare the performance of \method{} against three baselines for four tasks: circle-peg insertion with a 3mm clearance, circle-peg insertion with a 2mm clearance, USB cable plugging and drawer opening. \Cref{tab:main_results} shows the overall results for the in-distribution tasks. We observe that for the easiest task, circle-peg insertion with the highest clearance (3mm), \method{} and TactileConcat significantly outperform the vision-only and cross-attention models, achieving similar success rates. However, we notice that the percentage of direct insertions is considerably higher for our method, highlighting the model's ability to leverage tactile signals without degrading the vision ones. Further, both average maximum force and average completion time are lower for \method{}. This suggests that our modality fusion approach allows for time-efficient task execution as well as improved sensitivity to contact dynamics. This trend continues for the other three tasks, but we notice that the overall performance of the TactileConcat approach, when compared against \method{}, decreases as the tasks get harder. The insertion results can be visualized in \Cref{fig:main-plots}. 

\subsubsection{Out-of-Distribution Tasks}
In \Cref{tab:main_results}, we further evaluate the generalization capabilities of our model and the baseline approaches on out-of-distribution tasks, including square-peg insertion (2mm and 3mm), pentagon-peg insertion (2mm and 3mm) and HDMI cable plugging. For the higher-clearance insertion tasks, we observe results similar to the in-distribution setting, where \method{} and TactileConcat outperform the vision-only and cross-attention baselines. Our method again achieves the highest success rate and greatly improves the percentage of direct insertions, while also producing the lowest average maximum force and completion time. For the more challenging 2mm clearance insertion tasks, \method{} and TactileConcat methods exhibit comparable success rates. However, we observe a notable increase in force magnitude for TactileConcat, which exerts significantly higher forces relative to its in-distribution results. In contrast, \method{} maintains a low force and high direct insertion rate across its episodes. The HDMI cable plugging task further highlights these differences. While the vision-only and TactileConcat approaches exhibit near-zero percent success rates, \method{} significantly improves both success and direct insertion rates while maintaining a lower completion time. Average force values are comparable across methods. It is noteworthy that the cross-attention baseline consistently underperforms across all tasks, with only a slight relative improvement on the HDMI cable plugging task. We hypothesize that this behaviour is attributable to the limited amount of task-specific training data. Unlike \method{}, cross-attention introduces additional trainable interactions between visual and tactile representations, which may require substantially more data to learn effective multimodal correspondences.

Overall, these results indicate that FiLM conditioning yields more robust generalization and maintains sensitivity to contact dynamics under distribution shifts. In contrast, feature concatenation may be more susceptible to variations in contact geometry and exhibit lower sensitivity to contact dynamics, particularly in the 2mm clearance tasks. 

\begin{table*}[!htbp]
  \caption{\textbf{In-distribution and out-of-distribution evaluation results.} \method{} is compared with the baselines across diverse insertion and pulling tasks (30 rollouts per method for ID, 15 for OOD). We report success rates alongside safety-critical metrics such as maximum insertion force and the rate of direct insertions. \method{} consistently outperforms baselines in both reliability and execution efficiency while maintaining lower peak contact forces.}
  \label{tab:main_results}
  \centering
  \setlength{\tabcolsep}{6pt} 
  \scalebox{0.7}{ 
  \begin{tabular}{llllll}
    \toprule
    Task & Method & Success (\%) & Direct (\%) & Avg. Max Force (N) & Avg. Time (s) \\
    \midrule

    \rowcolor{gray!15}
    \multicolumn{6}{c}{\textit{In-Distribution Tasks}} \\
    \midrule

    \multirow{4}{*}{Circle-Peg (3mm)}
      & OpenVLA-OFT   & 86.67 & 3.33 & 14.94 {\scriptsize$\pm$ 4.66} & 92.24 {\scriptsize$\pm$ 48.10} \\
      & TactileConcat  & 96.67 & 16.67 & 9.19 {\scriptsize$\pm$ 3.45} & 75.11 {\scriptsize$\pm$ 37.28} \\
      & Cross-Attn & 63.33 & 10.00 & 8.16 {\scriptsize$\pm$ 7.18} & 155.13 {\scriptsize$\pm$ 31.59} \\
      & \method{} & \textbf{100.00} & \textbf{36.67} & \textbf{7.64} {\scriptsize$\pm$ 2.63} & \textbf{52.03} {\scriptsize$\pm$ 5.02} \\
    \midrule

    \multirow{4}{*}{Circle-Peg (2mm)}
      & OpenVLA-OFT   & 66.67 & \textbf{23.33} & 15.09 {\scriptsize$\pm$ 12.69} & 110.44 {\scriptsize$\pm$ 46.76} \\
      & TactileConcat  & 73.33 & 0.00 & 8.72 {\scriptsize$\pm$ 2.25} & 114.80 {\scriptsize$\pm$ 44.44} \\
      & Cross-Attn & 60.00 & 20.00 & 10.38 {\scriptsize$\pm$ 10.08} & 139.16 {\scriptsize$\pm$ 36.57} \\
      & \method{} & \textbf{86.67} & \textbf{23.33} & \textbf{7.22} {\scriptsize$\pm$ 2.00} & \textbf{87.11} {\scriptsize$\pm$ 37.59} \\
    \midrule

    \multirow{4}{*}{USB-Cable-Plug}
      & OpenVLA-OFT   & 33.33 & 0.00 & 15.01 {\scriptsize$\pm$ 9.09} & 164.52 {\scriptsize$\pm$ 27.06} \\
      & TactileConcat  & 43.33 & 6.67 & 12.96 {\scriptsize$\pm$ 5.04} & 135.11 {\scriptsize$\pm$ 56.82} \\
      & Cross-Attn & 33.33 & 0.00 & 26.23 {\scriptsize$\pm$ 14.98} & 134.58 {\scriptsize$\pm$ 52.96} \\
      & \method{} & \textbf{73.33} & \textbf{33.33} & \textbf{10.15} {\scriptsize$\pm$ 5.47} & \textbf{99.71} {\scriptsize$\pm$ 46.43}  \\
    \midrule

    \multirow{4}{*}{Open-Drawer}
      & OpenVLA-OFT   & 33.33 & 33.33 & 13.64 {\scriptsize$\pm$ 0.54} & 152.65 {\scriptsize$\pm$ 39.37} \\
      & TactileConcat  & 26.67 & 26.67 & \textbf{9.74} {\scriptsize$\pm$ 0.89} & 141.66 {\scriptsize$\pm$ 35.93} \\
      & Cross-Attn & 20.00 & 20.00 & 17.40 {\scriptsize$\pm$ 6.58} & 165.64 {\scriptsize$\pm$ 28.61} \\
      & \method{} & \textbf{86.67} & \textbf{73.33} & 10.84 {\scriptsize$\pm$ 1.87} & \textbf{94.33}              {\scriptsize$\pm$ 46.43}  \\
    \midrule
    
    \rowcolor{gray!15} & OpenVLA-OFT   & 58.10 & 12.38 & 14.94 {\scriptsize$\pm$ 9.16} & 126.72 {\scriptsize$\pm$ 51.34}\\
    \rowcolor{gray!15} & TactileConcat  & 64.76 & 10.48 & 10.27 {\scriptsize$\pm$ 4.12} & 113.04 {\scriptsize$\pm$ 52.31} \\
    \rowcolor{gray!15} & Cross-Attn  & 48.00 & 12.00 & 13.43 {\scriptsize$\pm$ 12.62} & 149.92 {\scriptsize$\pm$ 39.01} \\
    \rowcolor{gray!15} \multirow{-4}{*}{ {\strut Average}}  & \method{} & \textbf{86.67} & \textbf{37.14} & \textbf{8.65} {\scriptsize$\pm$ 3.80} & \textbf{81.72} {\scriptsize$\pm$ 38.00} \\
    \midrule
    \midrule

    \rowcolor{gray!15}
    \multicolumn{6}{c}{\textit{Out-of-Distribution Tasks}} \\
    \midrule
    \multicolumn{6}{c}{\textit{Circle-Peg (3mm)}} \\
    \midrule

    \multirow{4}{*}{Square-Peg (3mm)}
      & OpenVLA-OFT   & 93.33 & 0.00 & 18.31 {\scriptsize$\pm$ 8.84} & \textbf{51.60} {\scriptsize$\pm$ 5.62} \\
      & TactileConcat & 93.33 & 13.33 & 9.34 {\scriptsize$\pm$ 5.56} & 69.77 {\scriptsize$\pm$ 23.40}  \\
      & Cross-Attn & 60.00 & 6.67 & 6.49 {\scriptsize$\pm$ 1.14} & 165.27 {\scriptsize$\pm$ 18.74} \\
      & \method{}   & \textbf{100.00} & \textbf{46.67} & \textbf{5.37} {\scriptsize$\pm$ 0.41} & 52.95 {\scriptsize$\pm$ 4.68}  \\
    \midrule

    \multirow{4}{*}{Pentagon-Peg (3mm)}
      & OpenVLA-OFT   & 46.67 & 0.00 & 27.39 {\scriptsize$\pm$ 17.81} & 83.48 {\scriptsize$\pm$ 22.55}  \\
      & TactileConcat & \textbf{100.00} & 20.00 & 10.43 {\scriptsize$\pm$ 5.36} & 76.21 {\scriptsize$\pm$ 17.65} \\
      & Cross-Attn & 53.33 & 6.67 & 9.16 {\scriptsize$\pm$ 5.22} & 147.99 {\scriptsize$\pm$ 35.01} \\
      & \method{}   & \textbf{100.00} & \textbf{33.33} & \textbf{7.51} {\scriptsize$\pm$ 2.88} & \textbf{53.15} {\scriptsize$\pm$ 5.89} \\
    \midrule

    \multicolumn{6}{c}{\textit{Circle-Peg (2mm)}} \\
    \midrule

    \multirow{4}{*}{Square-Peg (2mm)}
      & OpenVLA-OFT   & 66.67 & 0.00 & 34.30 {\scriptsize$\pm$ 11.18} & \textbf{64.54} {\scriptsize$\pm$ 10.64}\\
      & TactileConcat & \textbf{86.67} & 6.67 & 27.72 {\scriptsize$\pm$ 8.36} & 91.83 {\scriptsize$\pm$ 10.64} \\
      & Cross-Attn & 53.33 & 6.67 & 26.24 {\scriptsize$\pm$ 12.36} & 156.33 {\scriptsize$\pm$ 28.07} \\
      & \method{}   & 80.00 & \textbf{40.00} & \textbf{7.06} {\scriptsize$\pm$ 1.36} & 111.61 {\scriptsize$\pm$ 38.31} \\
    \midrule

    \multirow{4}{*}{Pentagon-Peg (2mm)}
      & OpenVLA-OFT   & 60.00 & 0.00 & 21.62 {\scriptsize$\pm$ 17.87} & \textbf{80.91} {\scriptsize$\pm$ 18.05} \\
      & TactileConcat & 73.33 & 0.00 & 23.70 {\scriptsize$\pm$ 10.29}  & 116.56 {\scriptsize$\pm$ 30.55} \\
      & Cross-Attn & 46.67 & 6.67 & 20.65 {\scriptsize$\pm$ 14.01} & 145.28 {\scriptsize$\pm$ 34.90} \\
      & \method{}  & \textbf{86.67} & \textbf{20.00} & \textbf{10.50} {\scriptsize$\pm$ 7.72} & 104.61 {\scriptsize$\pm$ 37.60} \\
    \midrule

    \multicolumn{6}{c}{\textit{USB-Cable-Plug}} \\
    \midrule

    \multirow{4}{*}{HDMI-Cable-Plug}
      & OpenVLA-OFT   & 6.67 & 0.00 & \textbf{10.71} {\scriptsize $\pm$ 8.93} & 166.88 {\scriptsize$\pm$ 38.29} \\
      & TactileConcat  & 13.33 & 0.00 & 11.18 {\scriptsize$\pm$ 5.08} & 174.59 {\scriptsize$\pm$ 13.84} \\
      & Cross-Attn & 33.33 & 0.00 & 33.82 {\scriptsize$\pm$ 12.79} & 133.97 {\scriptsize$\pm$ 39.33} \\
      & \method{}   & \textbf{66.67} &\textbf{6.67} & 11.54 {\scriptsize$\pm$ 3.88} & \textbf{116.87} {\scriptsize$\pm$ 45.46} \\
    \midrule

    \rowcolor{gray!15} & OpenVLA-OFT   & 54.67 & 0.00 & 22.46 {\scriptsize$\pm$ 15.75} & 89.48 {\scriptsize$\pm$ 46.05} \\
    \rowcolor{gray!15} & TactileConcat & 73.33 & 8.00 & 16.47 {\scriptsize$\pm$ 10.54} & 105.79 {\scriptsize$\pm$ 43.16} \\
    \rowcolor{gray!15} & Cross-Attn  & 49.33 & 5.33 & 19.27 {\scriptsize$\pm$ 14.62} & 149.77 {\scriptsize$\pm$ 33.72} \\
    \rowcolor{gray!15} \multirow{-4}{*}{ {\strut Average}}  & \method{} & \textbf{86.67} & \textbf{29.33} & \textbf{8.40} {\scriptsize$\pm$ 4.71} & \textbf{87.84} {\scriptsize$\pm$ 42.69}\\
    \bottomrule
  \end{tabular}
  }
\end{table*}

\subsection{Ablation Study}
\begin{table*}[!htbp]
  \caption{\textbf{Ablation study and occlusion tests.} Top) Comparison of four FiLM integration locations within the vision backbone. Bottom) Robustness evaluation under varied camera conditions demonstrates that \method{} achieves the highest success rate despite visual interference. Methods are evaluated over 15 rollouts (240 total).}
  \label{tab:ablation}
  \centering
  \setlength{\tabcolsep}{6pt} 
  \scalebox{0.7}{  
  \begin{tabular}{clcccc}
    \toprule
    Task & Method & Success (\%) & Direct (\%) & Avg. Max Force (N) & Avg. Time (s) \\
    \midrule

    \rowcolor{gray!15}
    \multicolumn{6}{c}{\textit{FiLM Integration Stage}} \\
    \midrule

    \multirow{4}{*}{\shortstack{Circle-Peg (3mm)\\(In-Distribution)}}
      & AllFiLM  & \textbf{100.00} & 36.67 & 7.64 {\scriptsize$\pm$ 2.63} & \textbf{52.03} {\scriptsize$\pm$ 5.02} \\
      & EarlyFiLM   & 93.33 & \textbf{60.00} & \textbf{6.69} {\scriptsize$\pm$ 0.92} & 60.85 {\scriptsize$\pm$ 8.26} \\
      & MiddleFiLM  & \textbf{100.00} & 26.67 & 7.14 {\scriptsize$\pm$ 1.71} & 53.88 {\scriptsize$\pm$ 5.08} \\
      & LateFiLM & \textbf{100.00} & 23.33 & 8.47 {\scriptsize$\pm$ 2.71} & 54.74 {\scriptsize$\pm$ 7.34} \\
    \midrule

    \multirow{4}{*}{\shortstack{Pentagon-Peg (3mm)\\(Out-of-Distribution)}}
      & AllFiLM  & \textbf{100.00} & 33.33 & \textbf{7.51} {\scriptsize$\pm$ 2.88} & \textbf{53.15} {\scriptsize$\pm$ 5.89} \\
      & EarlyFiLM  & \textbf{100.00} & 33.33 & 9.96 {\scriptsize$\pm$ 4.49} & 77.40 {\scriptsize$\pm$ 21.68} \\ 
      & MiddleFiLM  & \textbf{100.00} & \textbf{53.33} & 8.99 {\scriptsize$\pm$ 4.49} & 53.57 {\scriptsize$\pm$ 6.89} \\
      & LateFiLM & \textbf{100.00} & 40.00 & 9.42 {\scriptsize$\pm$ 3.31} & 68.57 {\scriptsize$\pm$ 27.47} \\
    \midrule
    
    \rowcolor{gray!15}
    \multicolumn{6}{c}{\textit{Camera Condition}} \\
    \midrule

     \multirow{4}{*}{\shortstack{Circle-Peg (3mm)\\(80\% Dimmed)}}
      & OpenVLA-OFT & 93.33 & 0.00 & 16.29 $\pm$ 9.85 & 73.50 $\pm$ 8.41 \\
      & TactileConcat & 86.67 & 26.67 & 11.15 {\scriptsize$\pm$ 9.21} & 78.03 {\scriptsize$\pm$ 30.16} \\
      & Cross-Attn & 53.33 & 0.00 & 9.29 {\scriptsize$\pm$ 5.96} & 159.96 {\scriptsize$\pm$ 29.15} \\
      & \method{} & \textbf{100.00} & \textbf{26.67} & \textbf{8.62} {\scriptsize$\pm$ 2.13} & \textbf{67.79} {\scriptsize$\pm$ 6.25} \\
    \midrule

    \multirow{4}{*}{\shortstack{Circle-Peg (3mm)\\ (50\% Frames)}}
      & OpenVLA-OFT & 73.33 & 0.00 & 15.70 {\scriptsize$\pm$ 12.39} & 113.19 {\scriptsize$\pm$ 40.39} \\
      & TactileConcat & 80.00 & \textbf{40.00} & 14.64 {\scriptsize$\pm$ 13.91} & 71.52 {\scriptsize$\pm$ 34.51} \\
      & Cross-Attn & 46.67 & 0.00 & \textbf{7.53} {\scriptsize$\pm$ 1.25} & 161.10 {\scriptsize$\pm$ 30.70} \\
      & \method{} & \textbf{100.00} & 26.67 & 8.12 {\scriptsize$\pm$ 1.90} & \textbf{51.68} {\scriptsize$\pm$ 5.33} \\
    \bottomrule
  \end{tabular}
  }
\end{table*}

\subsubsection{FiLM Integration Stage} 
\label{sec:ablation}
By default, \method{} applies FiLM conditioning to all ViT blocks (denoted AllFiLM in \Cref{tab:ablation}). We ablate this choice to determine whether FiLM can be applied to only a subset of blocks. \Cref{tab:ablation} summarizes the results obtained when FiLM layers are introduced at different depths of the visual backbone (all, early, middle and late blocks). EarlyFiLM, MiddleFiLM and LateFiLM correspond to model variants where FiLM is integrated into a third of the ViT blocks at varying depths. We evaluate the fusion location results on the 3mm circle-peg insertion task and the OOD 3mm pentagon-peg task. Across all models, we observe comparable results, with high performance, suggesting that FiLM-based tactile conditioning is effective regardless of integration depth. The EarlyFiLM model, however, achieves a significantly higher percentage of direct insertions but yields comparable results on the other metrics. Despite this difference, all ablation variants perform similarly in the OOD generalization task. These findings suggest that applying FiLM conditioning to a limited number of ViT blocks suffices for tactile integration, keeping computational overhead low.

\subsubsection{Camera Condition}
We further conduct experiments to examine the effects of camera conditions on model performance and evaluate whether tactile readings can compensate for degraded visual inputs. These results can be visualized in the bottom portion of \Cref{tab:ablation}. Under dimmed lighting conditions (80\% reduction), \method{} maintains a 100\% success rate, outperforming both OpenVLA-OFT, TactileConcat and Cross-Attn. Additionally, our method achieves the lowest average maximum force and execution time. Under the second camera condition, a partially frozen stream (50\% frame updates), we notice a performance decrease for the vision-only baseline, whose success rate drops to 73.33\%. In contrast, \method{} again achieves 100\% success and maintains the lowest force and completion time of all baseline methods. Although TactileConcat achieves a higher percentage of direct insertions in this setting, it exerts a higher force magnitude and slower execution time relative to \method{}. For Cross-Attn, both camera conditions show decreased performance relative to other baselines as well as its performance on the regular circle-peg (3mm) task, demonstrating poor robustness to changes in camera condition. Overall, these results indicate that our approach improves robustness under visual degradation, allowing the policy to maintain performance when visual information is unreliable.

\subsubsection{Tactile Encoders Evaluation}
To select the tactile encoder for our main experiments, we evaluate several pretrained models on binary classification tasks drawn from T3 \cite{zhao2024transferable} and Sparsh \cite{higuera2024sparsh} benchmarks that isolate tactile signatures relevant to insertion. We assess the learned representations on binary classification tasks by training an MLP classifier on the tactile embeddings. These tasks were selected because they isolate tactile signatures characteristic of the contact-rich phases in insertion tasks. Tasks one and two, \textit{Rotation-High} and \textit{Rotation-Low}, evaluate the model's ability to detect whether the peg held by the gripper is rotated relative to its initial position, thereby capturing sensitivity to changes in contact geometry. Task three, \textit{Contact}, assesses whether the model can distinguish between contact and non-contact states on the peg, testing the encoder's ability to capture force and deformation-related features. To further test the pretrained models' force-encoding capacity, we evaluate their performance on the continuous force regression task from the Sparsh TacBench benchmark \cite{higuera2024sparsh}. \Cref{tab:pretrained_tactile_comparison} compares the state-of-the-art encoders across these four tasks. Based on the higher performance observed in these evaluations, we adopt Sparsh-DINO as the visuotactile backbone for all subsequent experiments.

\begin{table*}[!htbp]
  \caption{\textbf{Comparison of pretrained visuotactile representations.} Comparison of four pretrained tactile representations across three binary classification tasks and one TacBench task. }
  \label{tab:pretrained_tactile_comparison}
  \centering
  \setlength{\tabcolsep}{6pt} 
  \scalebox{0.8}{
  \begin{tabular}{lcccc}
    \toprule
    Dataset  & T3 & Sparsh-IJEPA & Sparsh-MAE & Sparsh-DINO \\
    \midrule

    \rowcolor{gray!15}
    \multicolumn{5}{c}{\textit{Classification Tasks}} \\
    \midrule
    
    Rotation-High (\%) & 92.73 & 99.15 & \textbf{99.36} & \textbf{99.36} \\
    Rotation-Low (\%)  & 83.09 & 96.44 & 96.64 & \textbf{98.42} \\
    Contact (\%)       & 73.31 & 85.08 & 93.92 & \textbf{95.39} \\
    \midrule
    Average & 83.04 & 93.56 & 96.64 & \textbf{97.72} \\
    \midrule
    \rowcolor{gray!15}
    \multicolumn{5}{c}{\textit{TacBench Task}} \\
    \midrule
    
    Force Estimation (RMSE) & 58.64 & 40.27 & 36.61 & \textbf{36.09} \\

    \bottomrule 
  \end{tabular}
  }
\end{table*}
\vspace{-20pt}

\section{Conclusion}
\begin{sloppypar}
In this work, we investigated the integration of tactile signals into VLA models via post-training finetuning approaches. We evaluated multiple pretrained tactile representations and demonstrated that they allow for effective visuotactile policy adaptation without requiring task-specific encoder training. We proposed \method{}, an image conditioning fusion method that uses feature-wise linear modulation to integrate tactile signals, and showed that it improves task performance, increases direct insertion rates, and reduces both completion time and interaction forces. Importantly, our empirical results indicate that feature-level tactile conditioning yields more stable generalization behaviour than concatenation- and cross-attention-based fusion as well as vision-only models.
\end{sloppypar}

\subsection{Limitations}
A broader evaluation of diverse manipulation tasks would further strengthen our results. 
However, without precise visuotactile simulators, extending the task set proved difficult, as the experiments had to be conducted in a real-world setup, which required extensive data collection and time-consuming rollouts. Additionally, while our FiLM-based fusion approach is designed around the OpenVLA-OFT architecture, extending it to other VLA backbones such as $\pi_{0.5}$ remains future work.

\maxpage{15}  %

\bibliographystyle{splncs04}
\bibliography{main}
\end{document}